\documentclass[lettersize,journal]{IEEEtran}
\usepackage{amsmath,amsfonts}
\usepackage{algorithmic}
\usepackage{algorithm}
\usepackage{array}
\usepackage[caption=false,font=normalsize,labelfont=sf,textfont=sf]{subfig}
\usepackage{booktabs}
\usepackage{makecell}
\usepackage{textcomp}
\usepackage{stfloats}
\usepackage{url}
\usepackage{verbatim}
\usepackage{graphicx}
\usepackage{cite}
\usepackage{hyperref}
\usepackage{cleveref}
\usepackage{orcidlink}
\usepackage{balance}
\usepackage{tikz}
\hypersetup{
    colorlinks=true,
    linkcolor=blue,
    filecolor=magenta,      
    urlcolor=cyan,
    pdfpagemode=FullScreen,
    }
\crefname{figure}{Fig.}{Figs.}
\crefname{equation}{Eq.}{Eqs.}

\begin{document}

\bstctlcite{IEEEexample:BSTcontrol}

\title{URF: A Unified Robot Control-Policy Framework \\for Stable Contact Aware Manipulation}

\author{
Jiyou Shin$^{1}$\orcidlink{0009-0007-8952-1644}, 
Youngjin Seo$^{2}$\orcidlink{0009-0004-8247-6340}, 
Jaeseog Won$^{2}$\orcidlink{0009-0003-9504-5059}, 
Sungwon Seo$^{1}$\orcidlink{0000-0001-7862-7936}, \\
Hyunjun Kim$^{2}$\orcidlink{0009-0005-6260-8059},
Seokmin Yoon$^{2}$\orcidlink{0009-0006-4744-8656},
Tuan Luong$^{1}$,\orcidlink{0000-0001-5490-4337}
and Hyungpil Moon$^{1,2*}$\orcidlink{0000-0002-1091-0716}

\thanks{This work has been submitted to the IEEE for possible publication. Copyright may be transferred without notice, after which this version may no longer be accessible
This work was supported by the Technology Innovation Program funded by the Ministry of Trade, Industry and Energy (MOTIE, Korea) under Grant RS-2026-25449082 and Grant 2410012091 (00443027).
\textit{(Corresponding author: Hyungpil Moon.)}}
\thanks{
$1$: The authors are with the Faculty of Mechanical Engineering, Sungkyunkwan University, Suwon 2066, South Korea
(e-mail: danny384@g.skku.edu; ssw0536@g.skku.edu; luongtuan@g.skku.edu; hyungpil@skku.edu)
}
\thanks{
$2$: The authors are with the Faculty of Intelligent Robotics, Sungkyunkwan University, Suwon 2066, South Korea
(e-mail: soh2879@g.skku.edu; jaeseogwon@g.skku.edu; hyunjun.1104@g.skku.edu; yosmon@g.skku.edu; hyungpil@skku.edu)}
}

\markboth{IEEE Robotics and Automation Letters. Preprint Version. June, 2026}%
{Shin \MakeLowercase{\textit{et al.}}: Unified_Robot_Framework}

\maketitle

\begin{abstract}
Learning-based manipulation policies usually predict robot actions from sensory observations and leave their execution to a separate low-level controller.
In rigid contact, this separation can be problematic: the same motion to a virtual target or compliant motion command can lead to unstable contact, tracking error, excessive loading, or tool damage, depending on the low-level controller.
In this paper, we propose a \textit{Unified Robot Control-Policy Framework} (URF), which connects compliant action prediction with unified impedance-admittance control.
Given multimodal observations, URF predicts a virtual target, a stiffness matrix, and an impedance-admittance switch ratio.
The switch ratio determines when the controller should behave more like admittance control for accurate motion tracking and when it should move toward impedance control for safer rigid contact.
Because demonstration data do not provide ground-truth environment stiffness, we construct switch-ratio labels from measured contact forces and use them to supervise controller-mode prediction.
Across box-flipping and line-pressing tasks, URF achieves higher task success rates while reducing failure modes observed with admittance-only execution, including rapid force buildup, large force oscillations, tool breakage, and robot safety stops.
These results suggest that contact-aware policies benefit from predicting not only compliant actions but also the controller behavior used to execute them. 
\\Project page: \href{https://jiyou384.github.io/urf_project_page/}{urf-project-page}
\end{abstract}


\begin{figure*}[!t]
\centering
\includegraphics[width=\textwidth]{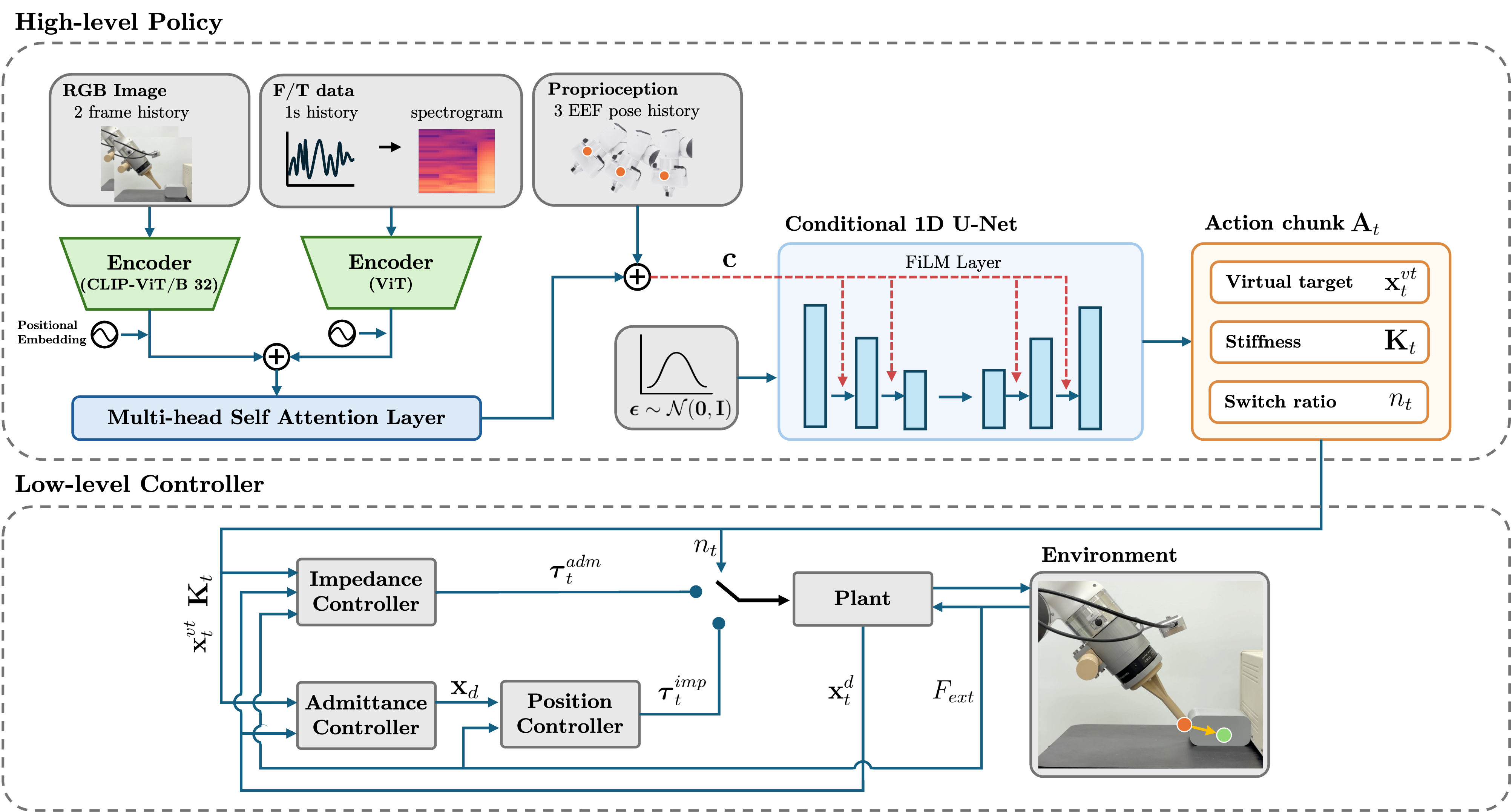}
\caption{Overview of the proposed URF network architecture and control-policy framework}
\label{fig:network_architecture}
\end{figure*}

\section{Introduction}
\IEEEPARstart{R}{ecent} advances in learning-based robot manipulation have enabled policies to generate complex action trajectories directly from sensory observations, such as images, joint angles, and end-effector poses~\cite{OpenVLA,RT2}.
In this paradigm, robot learning has shifted from hierarchical modular pipelines that decompose a task into separate perception, planning, and control modules to end-to-end policies that map observations directly to actions~\cite{visuomotor,ACT,DP,pi05,gr00t,geminirobotics}.
These approaches have shown impressive performance across a wide range of manipulation problems, driven by increasingly expressive architectures and generative models~\cite{Transformer,VAE,DDPM,FlowMatching}, and large-scale demonstration datasets~\cite{OpenX,DROID}.

Despite these advances, contact-aware manipulation remains challenging because task success depends on physical interaction as well as motion generation.
The robot must maintain contact, regulate force, and recover from small changes in the interaction with the object and the environment.
However, vision and proprioception alone may not provide sufficient information on contact state, such as contact onset, force buildup, or loss of contact~\cite{FACTR}.
To address this limitation, recent methods have incorporated contact force/torque or tactile measurements into observation~\cite{FoAR,ForceMimic,RDP}.
These additional sensory signals allow the policy to adapt when contact is made, lost, or unexpectedly loaded. 
Beyond using contact-related observations, natural direction is to make the policy action richer than a single reference pose.
Some methods~\cite{Admitdiff,TacDiffusion} predict interaction quantities, such as desired forces or 6D wrenches, and pass them to admittance control~\cite{adm_control1,adm_control2} or impedance control~\cite{ForceControl,ImpedanceControl}.
Others predict compliance-related quantities, including varying stiffness and virtual targets reconstructed from demonstrations~\cite{ACP,Comp-ACT}.
These action spaces allow the policy to express how contact should be handled, rather than only where the robot should move.

However, predicting contact-aware action alone does not determine how the predicted command will physically be executed.
In many policy-learning pipelines, the trained model predicts a target command, while a separately designed low-level controller converts that command into robot motion or torque commands~\cite{ACT,DP,OpenVLA}.
This separation overlooks the role of the low-level controller, even though it directly applies the predicted action to the physical interaction between the robot and the environment.
As a result, the same force, stiffness, or virtual target can produce different contact forces, tracking errors, or failure depending on the controller that receives it~\cite{Admitdiff,TacDiffusion,ACP,Comp-ACT}.
This issue is closely related to the classical trade-off between admittance and impedance control.
Admittance control is effective for accurate motion tracking in free space and soft contact, but is known to be sensitive to contact instability in stiff environments because the force feedback is converted into vibratory motion commands~\cite{unstable_admit1,unstable_admit2}.
Impedance control is more robust in stiff contact because it regulates the dynamic relationship between motion and force in a dissipative way, although its free-space tracking can be less precise~\cite{ForceControl,ImpedanceControl}.
A unified impedance-admittance controller was proposed to move between these two behaviors by switching the control mode~\cite{UnifiedController,rhee2023hybrid}.
By combining admittance control for accurate free-space motion with impedance control for stable interaction in stiff contact, this approach can exploit the advantages of both behaviors depending on the interaction condition.

Although this controller trade-off is important, it can be less visible in compliant settings, where the tool, object, or environment allows smoother contact interactions.
Many experiments lower the stiffness of the controller, attach soft material to the tool, or use deformable objects to make contact less abrupt.
These choices are practical and often necessary for safe data collection, but can also make the role of the low-level controller seem less important.
Therefore, we argue that contact-aware policy learning should not treat action prediction and low-level control as separate components.
Instead, the policy should predict not only compliant action parameters, such as virtual targets and stiffness, but also the controller behavior used to execute them.

We propose a \textit{Unified Robot Control-Policy Framework} (URF), in which the policy predicts virtual targets, stiffness, and an impedance-admittance switch ratio.
Instead of using a fixed low-level controller, URF uses the predicted switch ratio to adjust the controller mode during the task.
The switch ratio determines whether the low-level controller favors admittance-like tracking or impedance-like stable contact during execution.
Ideally, this ratio would be selected based on the environment stiffness~\cite{UnifiedController}, but this quantity is not available in demonstrations.
To train controller-mode prediction without environment stiffness data, we propose to use the measured force magnitude as a supervision signal for the impedance-admittance switch ratio.
This converts raw demonstration data into controller-aware action labels that indicate when the low-level controller should favor accurate motion tracking and when it should shift toward safer rigid-contact behavior.
Although the label is derived from a simple force-based rule, the trained policy is not a pure force-threshold controller: it predicts future switch ratios from images, proprioception, and force history, allowing it to reduce the switch ratio before large contact forces appear.
This provides a practical way to learn controller-aware actions from demonstrations while avoiding direct environment stiffness estimation.

We evaluate URF on two rigid contact-aware tasks: box-flipping with a 3D-printed tool and a stiff box, and line-pressing on a rigid surface.
These tasks test whether learned contact-aware actions can maintain stable contact under rigid physical interaction.
The results show that low-level control is critical for reliable contact-aware manipulation.

The main contributions of this work are as follows:
\begin{itemize}
\item {
\textbf{Controller-aware unified policy framework}. 
We introduce a control-policy framework that unifies learned action prediction and low-level interaction control for contact-aware manipulation.
By predicting virtual targets, stiffness, and an impedance-admittance switch ratio, URF adapts low-level control during execution and reduces contact-induced failures while preserving task-level motion tracking.
}
\item {
\textbf{Force-supervised impedance-admittance switching}. 
We address the lack of environment stiffness annotations in demonstrations by deriving impedance-admittance switch-ratio labels from the measured force magnitude.
This allows URF to learn when to favor admittance-dominant tracking or impedance-dominant stable contact from force-annotated demonstrations without explicit environment stiffness estimation.
}
\end{itemize}

The rest of this paper is organized as follows. 
~\Cref{sec:URF_framework} presents the proposed URF framework, including the low-level unified controller, controller-aware action representation, and demonstration label construction. 
~\Cref{sec:experiments} describes the experimental setup and evaluates URF on contact-rich manipulation tasks. 
Finally, ~\Cref{sec:conclusion} concludes the paper and suggests future works and limitations.


\section{Unified Robot Control-Policy Framework}
\label{sec:URF_framework}
URF consists of a low-level unified interaction controller and a high-level policy that predicts controller-aware actions. 
The low-level controller executes contact-aware motion by switching between impedance and admittance phases while receiving a time-varying virtual target, stiffness matrix, and switch ratio. 
The high-level policy generates these quantities from multimodal observations, including images, end-effector poses, and force/torque histories. 
This design allows the robot to adapt not only the desired contact interaction, represented by the virtual target and stiffness, but also how the interaction is controlled through the impedance-admittance switch ratio.

\subsection{Low-level Controller}
We use the unified impedance-admittance controller~\cite{UnifiedController} as the low-level interaction controller.
To apply it to contact-aware manipulation, we modify the controller to receive $\mathbf{x}^{vt}_t$, $\mathbf{K}_t$, and $n_t$ as time-varying inputs, where $\mathbf{x}^{vt}_t$ is the virtual target tracked by the controller, $\mathbf{K}_t$ is the stiffness matrix, and $n_t$ determines the duty cycle between the impedance and admittance modes.

In the impedance phase, the controller computes the impedance force command so that the end-effector follows the desired interaction dynamics around the virtual target. 
We define the virtual-target error as $\mathbf{e}_t = \mathbf{x}_t - \mathbf{x}^{vt}_t$ and $\dot{\mathbf{e}}_t = \dot{\mathbf{x}}_t - \dot{\mathbf{x}}^{vt}_t$, where $\mathbf{x}_t$ and $\dot{\mathbf{x}}_t$ are the end-effector position and velocity, respectively. 
The desired impedance behavior is given by
\begin{equation}
\mathbf{M}_d \ddot{\mathbf{e}}_t
+
\mathbf{D}_d \dot{\mathbf{e}}_t
+
\mathbf{K}_t \mathbf{e}_t
=
\mathbf{F}^{ext}_t ,
\label{eq:impednace}
\end{equation}
where $\mathbf{M}_d$ and $\mathbf{D}_d$ are the desired inertia and damping matrices, and $\mathbf{F}^{ext}_t$ is the measured external force. 
Substituting this desired interaction law into the robot Cartesian dynamics $\mathbf{M}_r \ddot{\mathbf{x}}_t = \mathbf{F}^{imp}_t + \mathbf{F}^{ext}_t$ yields the impedance force command in Cartesian coordinates
\begin{equation}
    \mathbf{F}^{imp}_t
    =
    \left(
    \mathbf{M}_r \mathbf{M}_d^{-1} - \mathbf{I}
    \right)
    \mathbf{F}^{ext}_t
    -
    \mathbf{M}_r \mathbf{M}_d^{-1}
    \left(
    \mathbf{D}_d \dot{\mathbf{e}}_t
    +
    \mathbf{K}_t \mathbf{e}_t
    \right),
    \label{eq:impedance}
\end{equation}
where $\mathbf{M}_r$ denotes the robot apparent Cartesian inertia and $\mathbf{I}$ is the identity matrix. 
The resulting Cartesian force command is converted into the impedance joint torque command $\boldsymbol{\tau}^{imp}_t$ through the Jacobian: $\boldsymbol{\tau}^{imp}_t  = \mathbf{J}^{\top}_t \mathbf{F}^{imp}_t$.
This phase is used to improve robustness under stiff contact.

In the admittance phase, the measured external force is converted into a desired motion command through the outer admittance dynamics,
\begin{equation}
    \mathbf{M}_d \ddot{\mathbf{x}}^{d}_t
    +
    \mathbf{D}_d
    \dot{\mathbf{x}}^{d}_t
    +
    \mathbf{K}_t
    \left(
    \mathbf{x}^{d}_t - \mathbf{x}^{vt}_t
    \right)
    =
    \mathbf{F}^{ext}_t ,
    \label{eq:admittance}
\end{equation}
where $\mathbf{x}^{d}_t$ is the desired position command generated by the admittance model. 
This desired position command is tracked by an inner position controller,
\begin{equation}
\mathbf{F}^{adm}_t
=
\mathbf{K}_p
\left(
\mathbf{x}^{d}_t - \mathbf{x}_t
\right)
-
\mathbf{K}_v \dot{\mathbf{x}}_t ,
\label{eq:admittance}
\end{equation}
where $\mathbf{K}_p$ and $\mathbf{K}_v$ are positive-definite proportional and derivative gain matrices.
Similar to impedance control, $\mathbf{F}^{adm}_t$ is converted into the admittance joint torque command $\boldsymbol{\tau}^{adm}_t$ through the Jacobian: $\boldsymbol{\tau}^{adm}_t  = \mathbf{J}^{\top}_t \mathbf{F}^{adm}_t$.
This phase is useful for accurately tracking motion commands in free space or weak contact.

We set $n_t = 1$ for admittance-only execution and $n_t = 0$ for impedance-only execution. 
For intermediate values, the controller switches between the two phases within each control period, where $n_t$ determines the fraction of the period spent in the admittance phase. 
Following the unified controller formulation~\cite{UnifiedController}, the transition between the two modes is handled by state mapping. 
When switching from impedance to admittance control, the admittance states $\mathbf{x}^{d}_t$ and $\dot{\mathbf{x}}^{d}_t$ are initialized so that the commanded Cartesian force and its time derivative remain continuous across the switch. 
When switching from admittance to impedance control, the controller returns to the impedance law defined around the virtual target $\mathbf{x}^{vt}_t$. 
This switching construction preserves the continuity of the robot state variables and has been shown to yield stable switching behavior. 
Because both $\mathbf{K}_t$ and $n_t$ vary over time, URF can adapt the interaction stiffness and the impedance-admittance behavior according to the task phase instead of using a fixed interaction controller throughout the trajectory.

\subsection{High-level Policy}
The high-level policy predicts a sequence of controller-aware actions from multimodal observations. 
As shown in~\Cref{fig:network_architecture}, the policy consists of a multimodal observation encoder and a conditional diffusion policy~\cite{DP} that generates action chunks for the low-level unified controller~\cite{UnifiedController}.
\subsubsection{Observation Encoding}
At each query, the policy receives visual, proprioceptive, and force observations. 
The input consists of an observation history containing two RGB frames from a single camera, three end-effector poses, and one second of force/torque measurements.
Each RGB image is encoded by a pretrained CLIP ViT-B/32~\cite{CLIP} backbone into a 768-dimensional visual feature. 
The force history is converted into a six-channel log-spectrogram and encoded into a 768-dimensional force feature by a ViT encoder~\cite{ViT}. 
These features are fused using a Transformer encoder~\cite{Transformer} and concatenated with the proprioceptive state to form a global observation condition $\mathbf{c}$.

\subsubsection{Controller-Aware Action Representation}
Given the observation condition, a conditional 1D U-Net~\cite{U-Net} predicts a future action chunk~\cite{DP} consisting of 16 action steps.
During denoising, $\mathbf{c}$ is used to condition the U-Net through FiLM layers~\cite{film}, which modulate the intermediate feature maps and guide the generated action sequence.
Unlike conventional manipulation policies that predict only motion commands~\cite{ACT,DP}, our policy directly predicts the parameters required by the low-level controller:
\begin{equation}
    \label{eq:action}
    \mathbf{a}_t = \left[\mathbf{x}^{vt}_t, \mathbf{K}_t, n_t\right],
\end{equation}
where $\mathbf{x}^{vt}_t$ is the virtual target pose, $\mathbf{K}_t$ is the stiffness matrix, and $n_t \in [0, 1]$ is the impedance-admittance switch ratio.
Here, the virtual target $\mathbf{x}^{vt}_t$ is not treated as a pose that must be tracked exactly; instead, it defines the equilibrium pose of the impedance relationship, allowing the low-level controller to generate contact interaction based on the displacement between the robot state and the virtual target.
When the robot is in contact, the error between the end-effector pose and $\mathbf{x}^{vt}_t$ is converted into an interaction force through the stiffness matrix $\mathbf{K}_t$.
In this action space, $\mathbf{x}^{vt}_t$ and $\mathbf{K}_t$ describe the intended contact behavior, while $n_t$ specifies the interaction control mode used for that behavior.

\subsubsection{Demonstration Label Construction}
We first collect a demonstration dataset through direct teaching, where a human operator physically guides the robot while visual observations, proprioceptive states, and force/torque measurements are recorded.
Given this dataset, we generate virtual target and stiffness labels using the adaptive compliance labeling procedure~\cite{ACP}.
The procedure uses the measured motion and force data to derive an approximate stiffness matrix $\mathbf{K}_t$ and reconstruct the corresponding virtual target trajectory $\mathbf{x}^{vt}_t$.

We also label the switch ratio $n_t$ for each demonstration sample. 
The unified-controller analysis suggests that the switch ratio should be adapted according to environment stiffness~\cite{UnifiedController}.
However, estimating the environment stiffness from demonstration data is highly challenging because it depends on unobserved contact geometry, material properties, and interaction dynamics.
We therefore use the measured force magnitude as a proxy for contact loading, as defined in~\Cref{eq:n}. 
Here, $f_{\min}$ and $f_{\max}$ are user-defined lower and upper force bounds that specify the force range over which the switch ratio changes.
Force magnitudes below $f_{\min}$ are assigned high switch ratios, corresponding to admittance-dominant execution for accurate motion tracking.
Force magnitudes above $f_{\max}$ are assigned low switch ratios, corresponding to impedance-dominant execution for stable contact under high loading force.
This labeling does not assume that large force necessarily implies high environment stiffness or that small force necessarily implies a compliant environment.
However, when environment stiffness is unavailable, large contact force should be handled conservatively because admittance-dominant execution can be unsafe under strong contact force.
For simplicity, we therefore use force magnitude as an indicator for shifting the controller toward impedance-dominant behavior, while keeping low-force regions admittance-dominant to preserve motion-tracking accuracy.
The experiments show that this force-based labeling is effective for improving contact stability without sacrificing task performance.
\begin{equation}
    \label{eq:n}
    n_t =
    \begin{cases}
    1 & |\mathbf{F}^{ext}_t| \leq f_{\min} \\
    1 - \dfrac{|\mathbf{F}^{ext}_t| - f_{\min}}{f_{\max} - f_{\min}}, & f_{\min} < |\mathbf{F}^{ext}_t| < f_{\max}, \\
    0, & |\mathbf{F}^{ext}_t| \geq f_{\max}.
    \end{cases}
\end{equation}
Consequently, low-force regions are labeled as admittance-dominant ($n_t \approx 1$), while high-force regions are labeled as impedance-dominant ($n_t \approx 0$). 
Intermediate force levels produce a smooth transition between the two behaviors.

\subsubsection{Policy Training and Execution}
We define an action chunk over the prediction horizon as $\mathbf{A}_t = [\mathbf{a}_t, \mathbf{a}_{t+1}, \dots, \mathbf{a}_{t+H-1}]$, where $H = 16$ and each per-step action $\mathbf{a}_t$ is defined in~\Cref{eq:action}.
The diffusion model is trained to generate this action chunk conditioned on the observation condition $\mathbf{c}$.
We train the policy with the diffusion noise-prediction loss~\cite{DDPM}.
For each ground-truth action chunk $\mathbf{A}_t^{0}$, we sample a diffusion timestep $\tau$ and Gaussian noise $\boldsymbol{\epsilon}$, corrupt the chunk to obtain $\mathbf{A}_t^{\tau}$, and train the conditional U-Net~\cite{U-Net} to predict the added noise from $\mathbf{A}_t^{\tau}$ and the observation condition $\mathbf{c}$.
\begin{equation}
    \label{eq:loss}
    \mathcal{L} =
    \mathbb{E}_{\mathbf{A}_t^{0},\,\tau,\,\boldsymbol{\epsilon}}
    \left[
    \left\|
    \boldsymbol{\epsilon}
    -
    \boldsymbol{\epsilon}_{\theta}
    \left(
    \mathbf{A}_t^{\tau},\,\tau,\,\mathbf{c}
    \right)
    \right\|_2^2
    \right]
\end{equation}
The virtual target, stiffness, and switch ratio at every step of the chunk are optimized jointly using the loss in~\Cref{eq:loss}.

At inference time, DDIM sampling~\cite{DDIM} produces an action chunk $\mathbf{A}_t$.
We execute its per-step actions $\mathbf{a}_t$, each comprising $\mathbf{x}^{vt}_t$, $\mathbf{K}_t$, and $n_t$, in a receding-horizon loop, re-predicting the next chunk from the latest visual, proprioceptive, and force observations.


\begin{figure}[t!] 
    \centering
    \includegraphics[width=0.9\columnwidth]{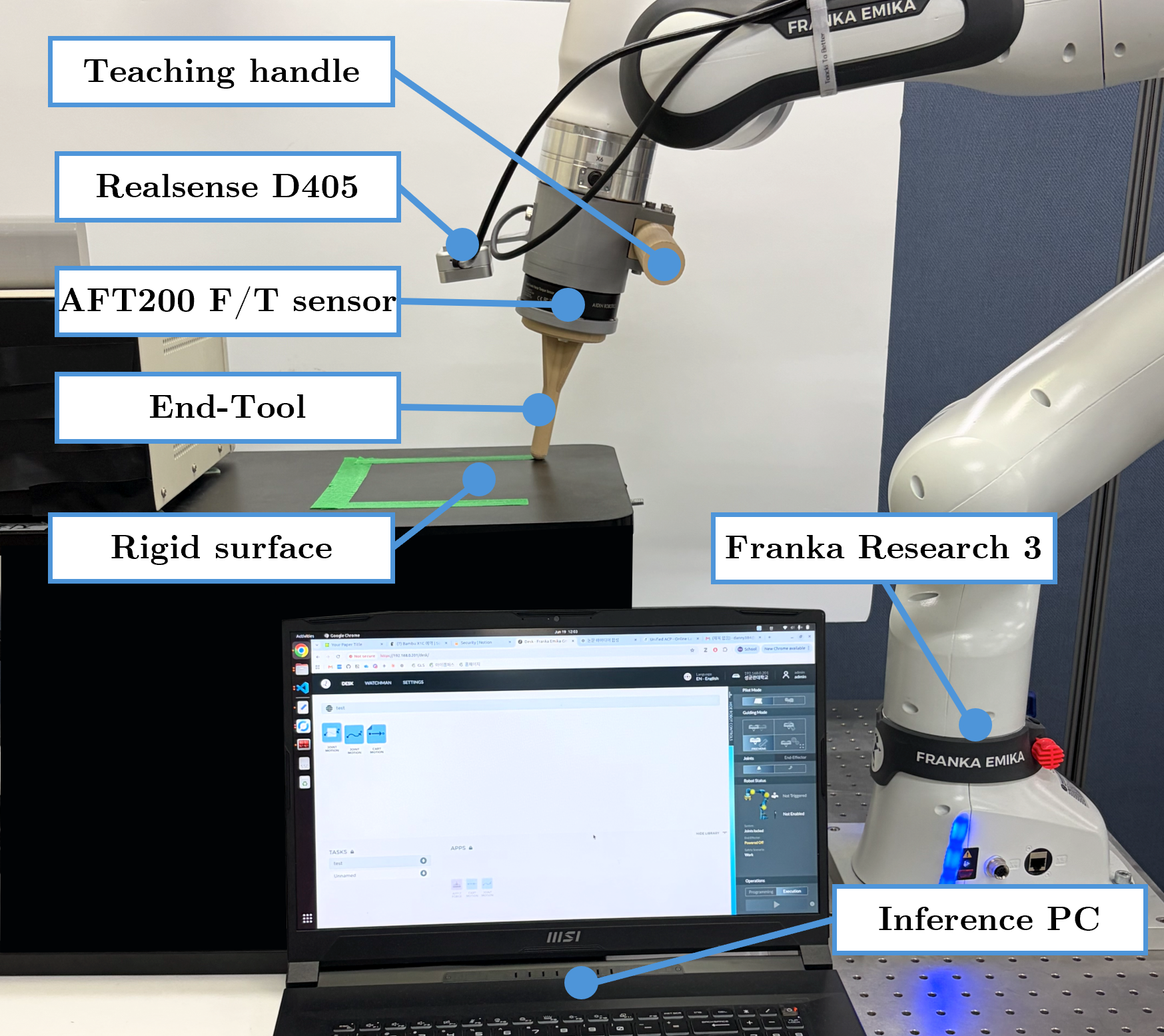}
    \caption{
    Experimental setup for the contact-aware manipulation tasks. 
    The setup includes rigid contact conditions used to evaluate the proposed controller-aware policy.
    }
    \label{fig:experiment_setup}
\end{figure}

\begin{figure*}[!t]
\centering
    \subfloat[]{
        \includegraphics[width=0.9\textwidth]{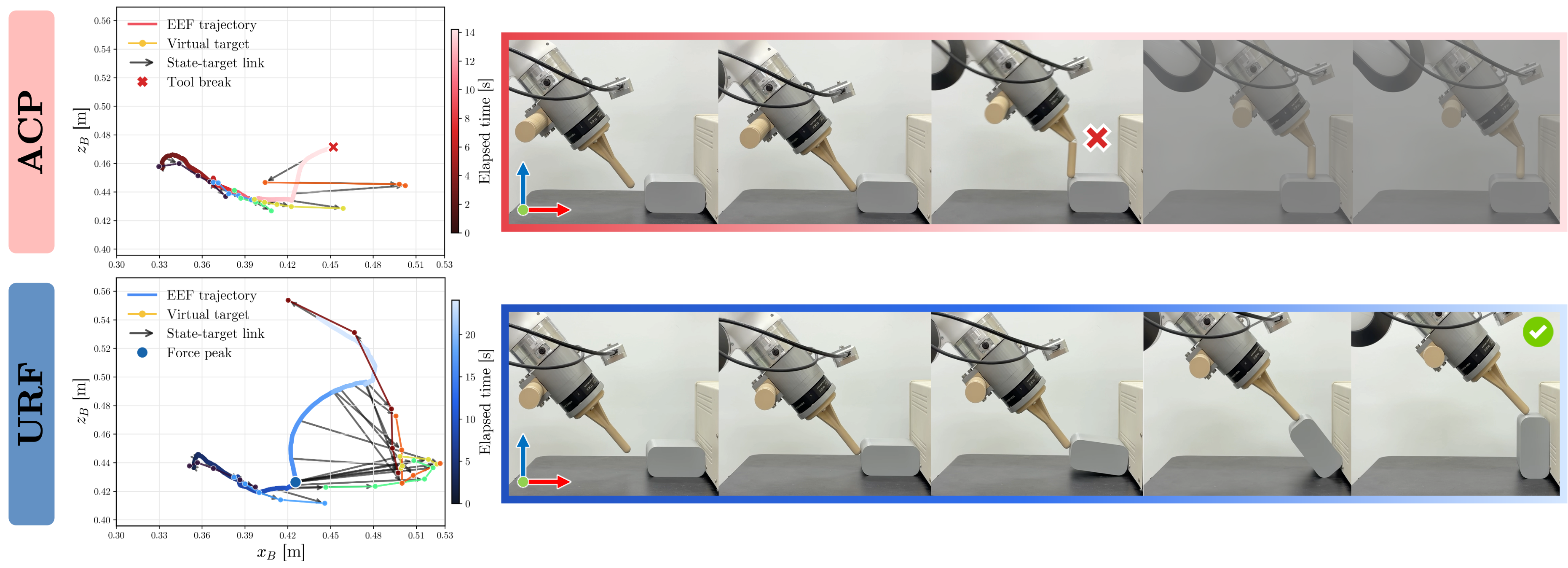}
        \label{fig:main_result_a}
    }
    \vspace{-2mm} \\
    \subfloat[]{
        \includegraphics[width=0.95\textwidth]{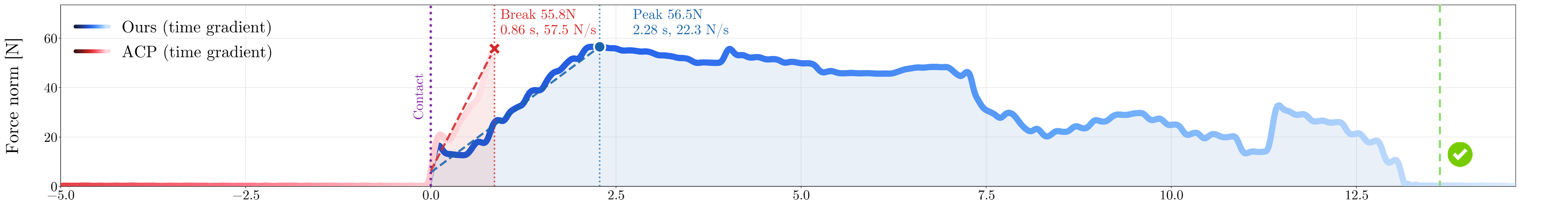}
        \label{fig:main_result_b}
    }
    \vspace{-2mm} \\
    \subfloat[]{
        \includegraphics[width=0.95\textwidth]{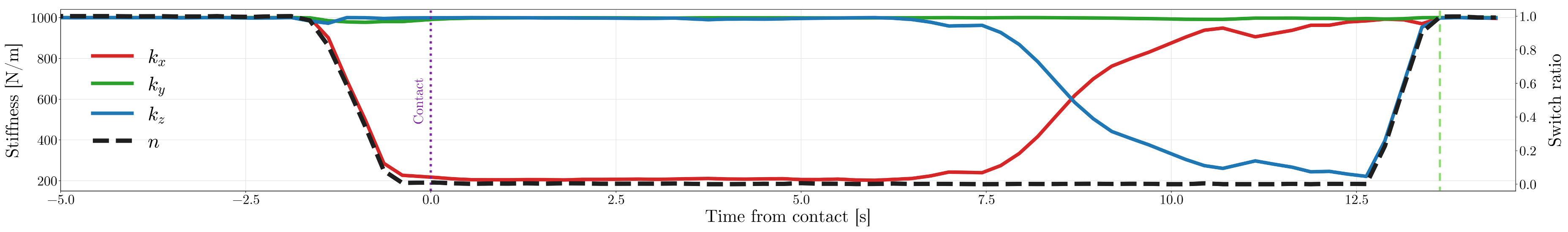}
        \label{fig:main_result_c}
    }
\caption{
Representative comparison between ACP and URF in the box-flipping task.
(a) The virtual target, actual end-effector trajectories in the $x$-$z$ plane and snapshots. 
Time progression is encoded by a dark-to-light trajectory colormap. 
Inferred horizons are visualized with matched colors, and arrows indicate the correspondence between each robot state and the virtual target commanded at that time. 
(b) The force norm of URF and ACP with the initial contact time aligned at $t=0$, using the same colormap as the end-effector trajectories to compare force increase and position at matched times.
ACP fails due to end-tool breakage $0.86\,\mathrm{s}$ after contact, whereas URF completes the task.
(c) The stiffness and switch-ratio predicted by URF, aligned at the contact instant. 
URF lowers $k_x$ and $n$ before contact, and the crossing of $k_x$ and $k_z$ during lifting marks the shift from pushing into the box to supporting the upward flipping motion.
}
\label{fig:main_result}
\end{figure*}

\section{Experiments}
\label{sec:experiments}
\subsection{Experiment Setup}
We evaluate the proposed method on two contact-aware manipulation tasks: box-flipping and line-pressing. 
Demonstrations are collected by direct teaching, where a human operator physically guides the robot with using a handle while visual observations, proprioceptive states, and force/torque measurements are recorded.
For the box-flipping task, we collect 100 demonstrations; a trial is successful when the robot fully lifts and rotates the box into an upright standing configuration. 
For the line-pressing task, we collect 50 demonstrations; a trial is successful only if the robot completes at least 85\% of the line while maintaining at least $5\mathrm{N}$ of contact force throughout the contact phase.
For each task and method, we conduct 20 evaluation trials.
We focus on rigid-contact settings because they expose the effect of low-level controller selection more clearly.
In softer settings, additional compliance from the tool, object, or environment can reduce contact-induced failures, and admittance-based compliance policies have already been shown to perform effectively~\cite{ACP}.
Therefore, our experiments use rigid tools, rigid objects, and rigid surfaces to evaluate whether controller-aware action prediction improves stability and task completion when contact forces are not passively absorbed by the environment.
The initial object poses are varied across trials to evaluate robustness to task-level variations.

We use a Franka Research 3 manipulator equipped with a wrist-mounted Intel RealSense D405 RGB camera and an AIDIN Robotics AFT200-D80 force/torque sensor mounted on the end flange with end-effector load and crosstalk compensation~\cite{FTcalibration}.
The RGB camera streams images at $60\,\mathrm{Hz}$, and the force/torque sensor provides raw six-axis wrench measurements at $1,000\,\mathrm{Hz}$, which are low-pass filtered and downsampled to $100\,\mathrm{Hz}$ for data collection and policy input.
Both the end tool and the box used in the flipping task are 3D-printed from PLA to create a rigid-contact manipulation setting.
Policy inference runs on an external PC with an NVIDIA RTX 4060 GPU. 
The predicted virtual target, stiffness, and switch ratio are transmitted to a real-time control PC through Franka ROS2. 
The low-level controller runs at $1,000\,\mathrm{Hz}$ and sends torque commands to the robot.
The overall hardware and task setup is shown in~\Cref{fig:experiment_setup}.

We compare URF with the following baselines:
\begin{itemize}
\item {
\textbf{DP w/ force}: A standard diffusion policy~\cite{DP} baseline with the same force encoding as URF. 
It does not predict virtual targets or compliance commands, and its actions are executed by a critically damped PD position controller with a translational stiffness of $k = 600\,\mathrm{N/m}$.}
\item {
\textbf{ACP}: Adaptive Compliance Policy~\cite{ACP}, which predicts virtual targets, adaptive stiffness and executes them using the admittance controller.
In the notation of our unified controller, this corresponds to the full-admittance execution setting.}
\item {
\textbf{URF w/} $n=0.5$: An ablation of URF that uses the proposed unified controller but fixes the impedance-admittance switch ratio to $n=0.5$.}
\item {
\textbf{URF w/ $n=0$}: An ablation of URF that fixes the switch ratio to $n=0$, corresponding to impedance-dominant execution.}
\item {
\textbf{URF}: The full proposed method, which predicts the virtual target, stiffness, and switch ratio, enabling adaptive impedance-admittance control according to the task phase and contact condition.}
\end{itemize}


\begin{table}[!t]
\centering
\caption{Comparison with baselines and ablations \\on the box-flipping task}
\renewcommand{\arraystretch}{1.2}
\setlength{\tabcolsep}{3pt}
\label{tab:box_result}
\begin{tabular}{l|cccc}
\Xhline{3\arrayrulewidth}
Method & SR [\%] $\uparrow$ & CFR [\%] $\downarrow$ & PFGR [N/s] $\downarrow$ & FSTE [mm$^2$] $\downarrow$  \\ 
\hline
DP w/ force~\cite{DP}    & 0           & --      & --                & -- \\
ACP~\cite{ACP}  & 25          & 86.7    &  $48.5 \pm 15.5$  & $4.36 \pm 2.00$  \\
URF w/ $n=0.5$  & 60          & 0       &  $24.4 \pm 10.9$  & $5.98 \pm 1.46$  \\
URF w/ $n=0$    & 50          & 0       & $21.7 \pm 7.5$    & $6.47 \pm 2.15$  \\
URF (Ours)           & \textbf{90} & 0       & $23.1 \pm 5.7$    & $5.93 \pm 1.68$  \\
\Xhline{3\arrayrulewidth}
\end{tabular}
\begin{flushleft}
\footnotesize
SR: success rate, PFGR: peak force growth rate, FSTE: free-space tracking error, CFR: critical failure rate.
\end{flushleft}
\vspace{10pt}
\end{table}

\subsection{Results}
\subsubsection{Box-Flipping}
\Cref{fig:main_result} shows a representative result of the box-flipping task. 
With ACP, which uses an admittance controller, the end-effector tool breaks shortly after contact. 
Although the policy predicts a meaningful virtual target for generating contact force, the admittance controller continuously follows the inward virtual-target command during rigid contact, causing the contact force to increase rapidly as shown in~\Cref{fig:main_result_b}. 
In this trial, ACP reaches a peak force similar to that of URF, but its contact force grows with a much steeper slope of $57.5\,\mathrm{N/s}$, approximately twice that of URF as shown in~\Cref{fig:main_result_b}. 
This result suggests that failure is not explained by peak force alone; rather, the rate at which contact force builds up after contact plays a critical role in tool damage.

In contrast, URF maintains stable contact and successfully completes the box-flipping motion.
During the initial contact phase, the predicted virtual target moves in the positive $x$-direction, toward the inside of the object, to establish contact force. 
It then moves upward in the positive $z$-direction as the box starts to rotate. 
Before contact, the policy has already set $n \approx 0$ and reduced the contact-direction stiffness to its minimum value, $k_x\approx 200\,\mathrm{N/m}$. 
As lifting continues, $k_x$ increases and $k_z$ decreases until the two curves cross, which coincides with the box reaching a partially upright posture. 
After this point, the controller no longer primarily pushes the box along the $x$-direction; instead, it presses and supports the motion along $z$. 
When the flip is complete, the policy restores an admittance-dominant setting ($n\approx1$) and raises $k_z\approx1,000\,\mathrm{N/m}$ for accurate post-contact tracking as shown in ~\Cref{fig:main_result_c}.

For the box-flipping study, we report success rate (SR), critical failure rate (CFR), peak force growth rate (PFGR), and free-space tracking error (FSTE).
SR is computed over 20 evaluation trials.
CFR is computed as the ratio of safety-critical failures, such as tool breakage or robot safety stops, to all failed trials; non-damaging failures, such as missed contact or poor tracking, are excluded from the numerator.
PFGR is the average slope of the force norm from the first contact to the peak force; we use it to capture rapid force buildup that can precede tool breakage or robot safety stops. 
FSTE is the mean-squared error between the end-effector pose and the virtual target before contact. 

As shown in~\Cref{tab:box_result}, URF achieves the highest success rate of 90\%.
DP~\cite{DP} fails in all 20 trials. 
Without a virtual target or compliance action, it must establish contact through reference actions inferred from vision and proprioception; in practice, it misses the box or fails to apply sufficient load to start the flip. 
ACP~\cite{ACP} succeeds in only 25\% of the trials and has a high CFR because most of its failures involve tool breakage or safety stops. 
Its PFGR, $48.5\,\mathrm{N/s}$, is roughly twice that of the URF variants, matching the force buildup observed in~\Cref{fig:main_result_b}.

The three URF variants have similar PFGR values, which suggesting that impedance-like behavior near contact is sufficient to slow the initial force rise.
This does not, however, guarantee task success. 
With a fixed $n=0$ or $n=0.5$, the approach motion is less accurate, and several trials contact the box at a point that is too low for a clean flip. 
URF avoids this failure mode by using admittance-dominant tracking during the approach and lowering $n$ as rigid contact begins.
The FSTE values show the same trade-off: ACP~\cite{ACP} achieves the lowest free-space tracking error, URF w/ $n = 0$ has the highest error, and URF remains close to URF w/ $n = 0.5$ while achieving a much higher SR.
The ablation therefore suggests that box-flipping requires both stable force buildup at contact and accurate placement of the contact point, and URF satisfies these requirements by adapting the impedance-admittance behavior during execution.

\begin{figure}[t!] 
    \centering
    \subfloat[]{
        \includegraphics[width=0.9\columnwidth]{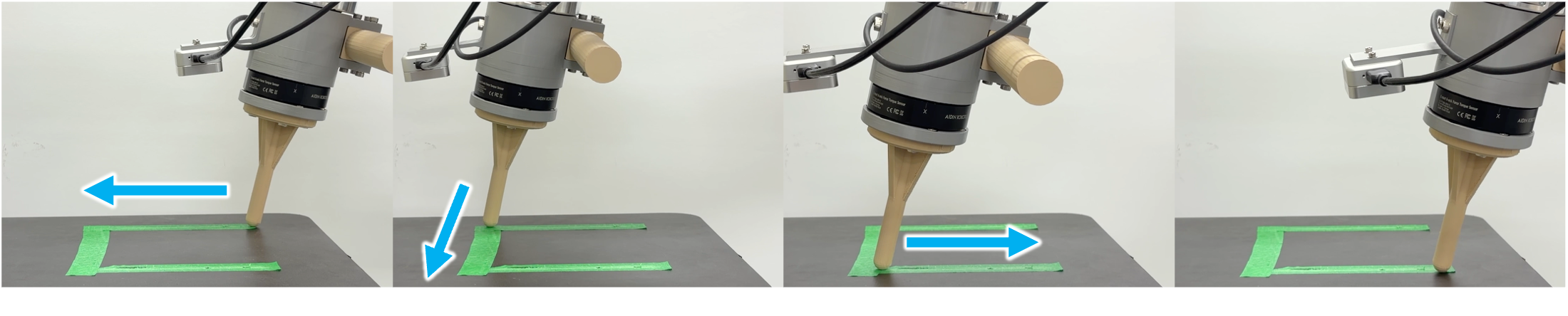}
        \label{fig:line_pressing_snapshots}
    }
    \vspace{-2mm} \\
    \subfloat[]{
        \includegraphics[width=0.9\columnwidth]{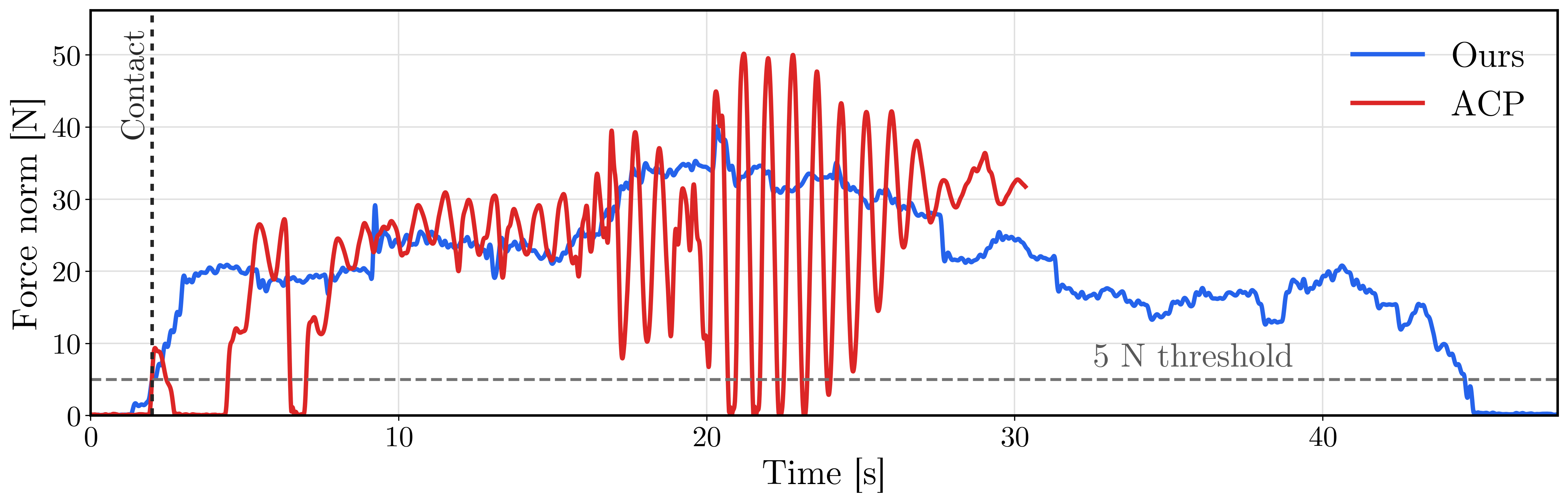}
        \label{fig:line_pressing_force_norm}
    }
    \vspace{-2mm} \\
    \subfloat[]{
        \includegraphics[width=0.9\columnwidth]{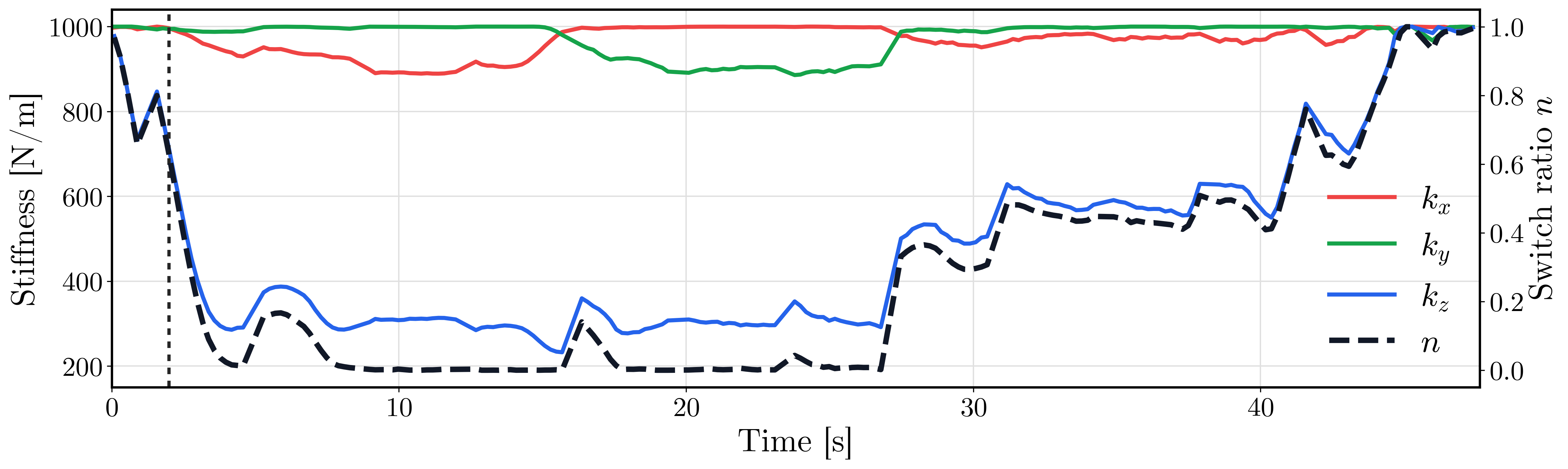}
        \label{fig:line_pressing_stiffness}
    }
    \caption{
    Representative results for the line-pressing task.
    (a) Snapshots of the task execution, where the robot follows a line on a rigid surface while applying contact force.
    (b) The second row compares the force norm of URF and ACP.
    URF maintains a stable contact force above $5\,\mathrm{N}$ after contact and completes the task, whereas ACP fails to establish stable contact at the beginning and later exhibits large force oscillations before triggering a robot safety stop.
    (c) The stiffness and switch-ratio profiles of URF.
    After contact, URF lowers the $k_z$ and $n$ to suppress force oscillations, then adjusts $n$ during motion so that the tool can maintain contact while staying on the line.}
    \label{fig:line_following_result}
\end{figure}

\subsubsection{Line-Pressing}
The line-pressing task evaluates sustained contact rather than a short flipping contact. 
The robot must follow the line while keeping the contact force above $5\,\mathrm{N}$ as shown in~\Cref{fig:line_pressing_snapshots}. 
We report contact-maintenance rate (CMR), the ratio of episodes that maintain contact throughout the task, and oscillation root-mean-square (ORMS) along with SR and CFR. 
ORMS is computed after subtracting a $1\,\mathrm{s}$ moving-average from the force norm, where lower values indicate smoother contact.

ACP~\cite{ACP} does not complete any line-pressing trial, and every failure ends with a robot safety stop, resulting in a CFR of $100\%$. 
In force norm plot in~\Cref{fig:line_following_result}, the force trace explains this failure mode. 
After the first contact, the force repeatedly drops near $0\,\mathrm{N}$ and rises toward $50\,\mathrm{N}$ before the robot stops as shown in~\Cref{fig:line_pressing_force_norm}. 
Because ACP~\cite{ACP} uses admittance-only execution, contact with the rigid surface becomes unstable, resulting in the highest ORMS among the evaluated methods. 
This suggests that the policy output alone is not sufficient to maintain a steady press when the low-level controller continuously converts force feedback into motion commands under rigid contact.

The URF variants behave differently as shown in ~\Cref{tab:line_result}. 
URF w/ $n=0$, URF w/ $n=0.5$, and URF all maintain contact in every trial, resulting in a CMR of $100\%$. 
They also achieve much lower ORMS than ACP~\cite{ACP}, indicating that impedance-like execution suppresses the large force oscillations observed with admittance-only control. 
In particular, URF w/ $n=0$ achieves the lowest ORMS because it remains impedance-dominant throughout the task. 
However, stable contact alone does not guarantee task success.
The fixed-switch variants can deviate from the line because of reduced tracking accuracy along the surface, causing some trials to fail before reaching the $85\%$ completion threshold.
URF succeeds in all trials because it adapts $n$ during pressing as shown in~\Cref{fig:line_following_result}.
It uses impedance-like behavior when stable contact is needed while retaining sufficient admittance-like behavior to follow the line accurately. 
As a result, URF satisfies both requirements of the task: continuous pressing above $5\,\mathrm{N}$ and sufficient progress along the line. 
This result highlights that minimizing force oscillation alone is insufficient; the controller must also preserve the tracking performance required by the task.

\begin{table}[!t]
\centering
\renewcommand{\arraystretch}{1.2}
\setlength{\tabcolsep}{3pt}
\caption{Comparison with baselines and ablations \\on the line-pressing task}
\label{tab:line_result}
\begin{tabular}{l|cccc}
\Xhline{3\arrayrulewidth}
Method & SR [\%] $\uparrow$ & CFR [\%] $\downarrow$& CMR [\%] $\uparrow$ & ORMS [N] $\downarrow$\\ 
\hline
DP w/ force~\cite{DP}    & 0            & --      & --   & -- \\
ACP~\cite{ACP}  & 0            & 100     & 0    & $3.22 \pm 0.81$ \\
URF w/ $n = 0.5$  & 70           & 0       & 100  & $1.20 \pm 0.28$ \\
URF w/ $n = 0$    & 70           & 0       & 100  & $0.92 \pm 0.24$ \\
URF (Ours)             & \textbf{100} & 0       & 100  & $0.96 \pm 0.36$ \\
\Xhline{3\arrayrulewidth}
\end{tabular}
\begin{flushleft}
\footnotesize
SR: success rate,
CMR: contact maintained rate,
CFR: critical failure rate,
ORMS: oscillation root-mean-square
\end{flushleft}
\vspace{10pt}
\end{table}

\begin{figure}[!t]
\centering
    \includegraphics[width=1.0\columnwidth]{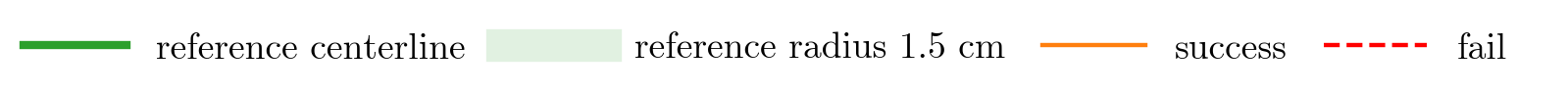}
    \\
    \vspace{-4mm}
    \subfloat[]{
        \includegraphics[width=0.49\columnwidth]{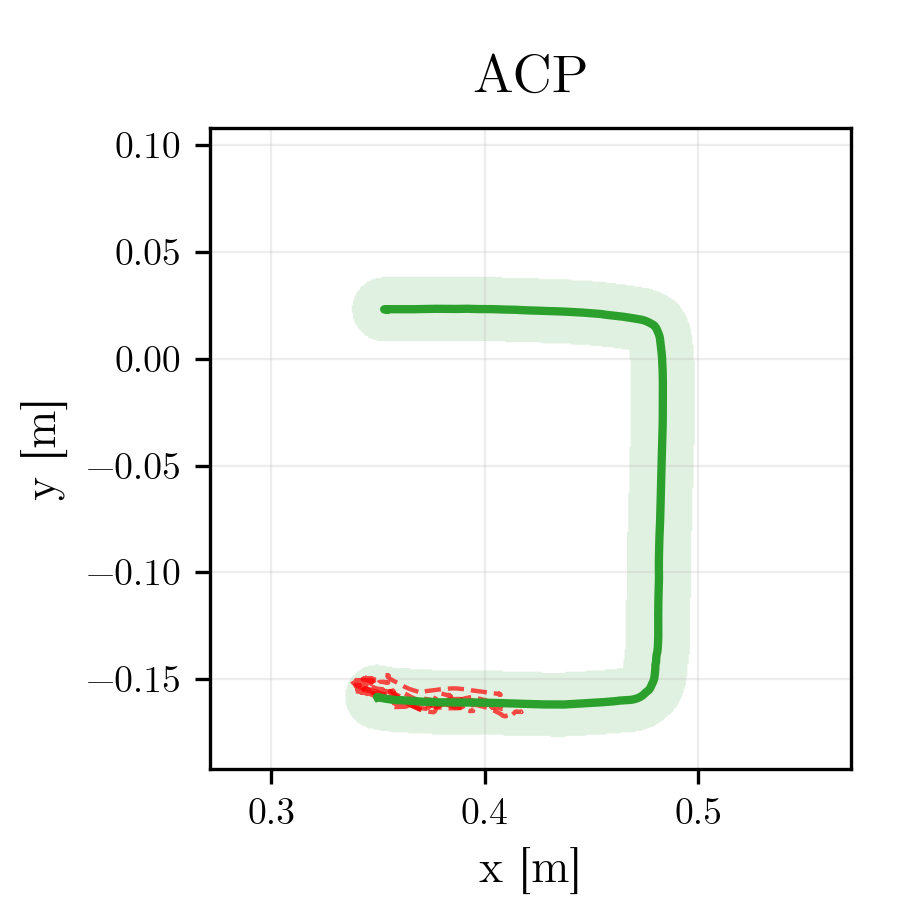}
        \label{fig:acp_line_traj}
    }
    \subfloat[]{
        \includegraphics[width=0.49\columnwidth]{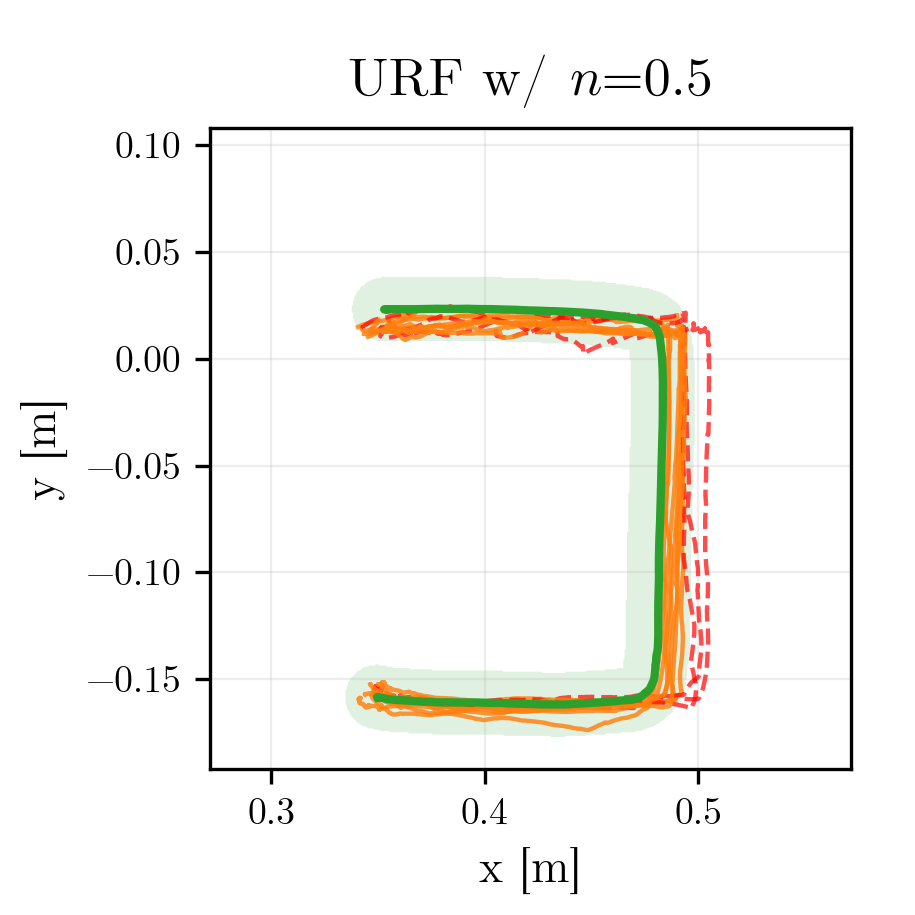}
        \label{fig:n0p5_line_traj}
    } \\
    \vspace{-3mm}
    \subfloat[]{
        \includegraphics[width=0.49\columnwidth]{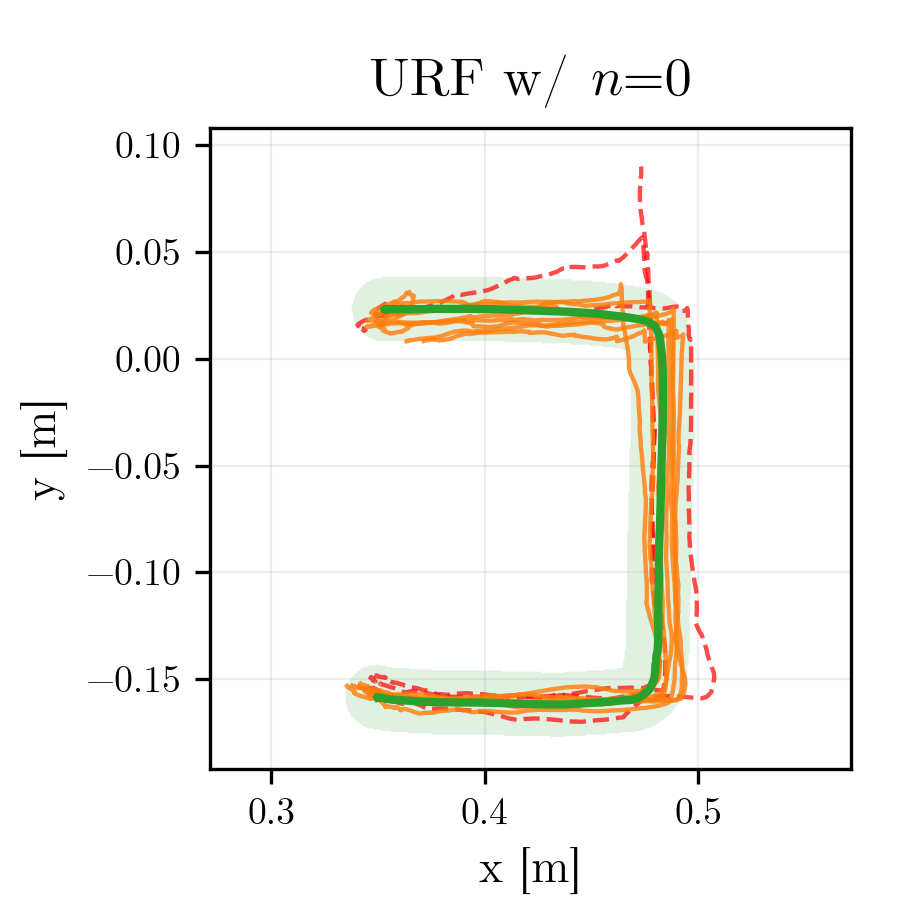}
        \label{fig:n0_line_traj}
    }
    \subfloat[]{
        \includegraphics[width=0.49\columnwidth]{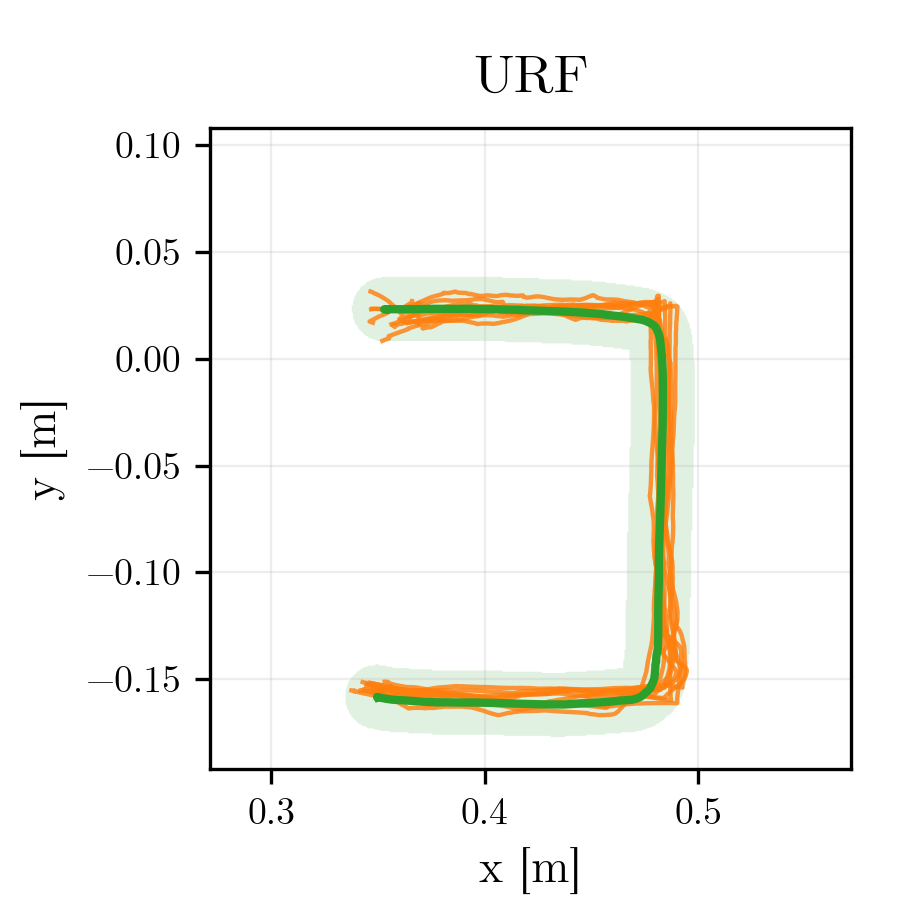}
        \label{fig:urf_line_traj}
    }
\caption{
Trajectory comparison in the line-pressing task.
Each plot shows ten evaluation trajectories for each method overlaid with the reference line.
(a) ACP fails during execution due to force oscillation.
(b) URF w/ $n=0.5$ follows the reference reasonably well, but some trajectories leave the valid region.
(c) URF w/ $n=0$ shows larger tracking error and higher trajectory variance due to impedance-only execution, with occasional large deviations from the reference.
(d) URF tracks the reference accurately and succeeds in all trials.
}
\label{fig:line_pressing_traj}
\end{figure}

\Cref{fig:line_pressing_traj} qualitatively compares the executed trajectories in the line-pressing task with the reference line.
Because the motion is performed under sustained contact, these trajectories should not be interpreted as pure free-space tracking accuracy.
As shown in~\Cref{fig:acp_line_traj}, ACP fails to maintain stable contact, and all trials terminate with a safety stop.
In~\Cref{fig:n0p5_line_traj}, URF w/ $n=0.5$ generally follows the reference, but some trajectories leave the valid region, causing task failures.
In~\Cref{fig:n0_line_traj}, URF w/ $n=0$ achieves the same success rate as URF w/ $n=0.5$, but shows larger trajectory variance and larger deviations in failed trials due to impedance-only execution.
In~\Cref{fig:urf_line_traj}, full URF keeps the trajectories tightly clustered around the reference and succeeds in all trials.
These results indicate that adaptive switching of $n_t$ improves path accuracy during pressing by using admittance-like tracking behavior while preserving stable contact.

\section{Conclusion}
\label{sec:conclusion}
We presented URF, a robot control-policy framework that predicts both compliant actions and the impedance-admittance mode used by the low-level controller. 
Existing contact-aware policies often treat action prediction and low-level control as separate components, which can make rigid-contact execution sensitive to the choice of controller.
URF addresses this problem by predicting a virtual target, stiffness, and switch ratio from multimodal observations.
The switch-ratio labels are obtained from the measured force magnitude, so the method does not require ground-truth environment stiffness.
In box-flipping, URF achieved the highest success rate and reduced the rapid force buildup that caused tool breakage in the admittance-only baseline. 
In line-pressing, URF kept contact force above the task threshold while following the line, whereas the admittance-only baseline produced large force oscillations and safety stops. 
The fixed-$n$ ablations showed that contact stability alone is insufficient: the robot also needs accurate approach and tangential tracking. 
These results suggest that, for reliable contact-aware manipulation across diverse interaction conditions, the learned policy and the low-level interaction controller should be designed as a coupled system rather than as separate components.

The current framework has two limitations.
First, the impedance-admittance switch ratio is shared across all control axes, even though more complex tasks, such as insertion, may require the robot to remain stiff along some axes while staying compliant along others.
Second, the force-supervised switch-ratio labels depend on user-defined force bounds, $f_{\min}$ and $f_{\max}$, whose appropriate values may vary across tasks, tools, and contact materials.

In future work, we plan to integrate the controller-aware action representation with vision-language-action models (VLA) or robotic foundation models (RFM).
Richer scene understanding could help infer task- and scene-dependent contact conditions, reducing the need for manually specified force bounds and supporting more complex contact-rich manipulation.
\balance
\bibliographystyle{IEEEtran}
\bibliography{reference}

\vfill

\end{document}